\documentclass[twoside,11pt]{article}

\usepackage{jmlr2e}

\usepackage{color}
\usepackage{amsmath,amssymb}
\usepackage{booktabs}
\usepackage{hyperref}
\hypersetup{
    colorlinks=true,
    linkcolor=red,          
    citecolor=blue,        
    filecolor=magenta,      
    urlcolor=blue,           
    pdfborder = {0 0 0}
 }

\usepackage{subfigure}
\usepackage{multirow}
\usepackage{xspace}

\newcommand\ie{\textit{i.e.}\xspace}

\renewcommand{\vec}[1]{\mathbf{\boldsymbol{#1}}}
\newcommand{\mat}[1]{\mathbf{\boldsymbol{#1}}}
\def\CC{{C\nolinebreak[4]\hspace{-.05em}\raisebox{.4ex}{\tiny\bf ++} \xspace}}

\jmlrheading{1}{2014}{1-38}{10/12}{\ldots}{Bj{\"o}rn Barz, Erik Rodner, and Joachim Denzler}

\ShortHeadings{ARTOS -- Adaptive Real-Time Object Detection System}{Barz, Rodner, Denzler}
\firstpageno{1}

\begin{document}


\title{ARTOS\\Adaptive Real-Time Object Detection System}

\author{\name Bj{\"o}rn Barz \email bjoern.barz@uni-jena.de \\
       \AND
       \name Erik Rodner \email erik.rodner@uni-jena.de \\
       \AND
       \name Joachim Denzler \email joachim.denzler@uni-jena.de \\[12pt]
       \addr Computer Vision Group\\
       Friedrich Schiller University of Jena\\
       Jena, Germany
       }

\editor{}

\maketitle

\begin{abstract}
ARTOS is all about creating, tuning, and applying object detection models with just a few clicks.
In particular, ARTOS facilitates learning of models for visual object detection by eliminating the burden of having to collect and annotate a large set of positive and negative samples manually and in addition it implements a fast learning technique to reduce the time needed for the learning step. 
A clean and friendly GUI guides the user through the process of model creation, adaptation of learned models to different domains using in-situ images, and object detection on both offline images and images from a video stream.
A library written in \CC provides the main functionality of ARTOS with a C-style procedural interface, so that it can be easily integrated with any other project.
\end{abstract}


\section{How does ARTOS work?}

Object detection is often one of the basic algorithms necessary for a lot of vision applications in
robotics or related fields. Our work is based on the ideas of several state-of-the-art papers and the setup described in \cite{Goehring14:ITR}. Therefore, we do not claim
any novelty in terms of methodology, but rather present an open source project that aims at making object detection learned on large-scale datasets available
to a broader audience. However, we also extend the system of \cite{Goehring14:ITR} with respect to the following aspects:
\begin{enumerate}
\item Multiple components for each detector similar by performing clustering
\item Threshold optimization using leave-one-out cross-validation and optimization of the mixture's threshold combination
\item Flexible interactive model tuning, \ie a user can \textit{remove} components from a model and add \textit{multiple} new models using in-situ images (images of the application environment)
\end{enumerate}

\paragraph{Sample Acquisition}

\begin{figure}[tbp]
    \centering
    \includegraphics[width=0.5\linewidth]{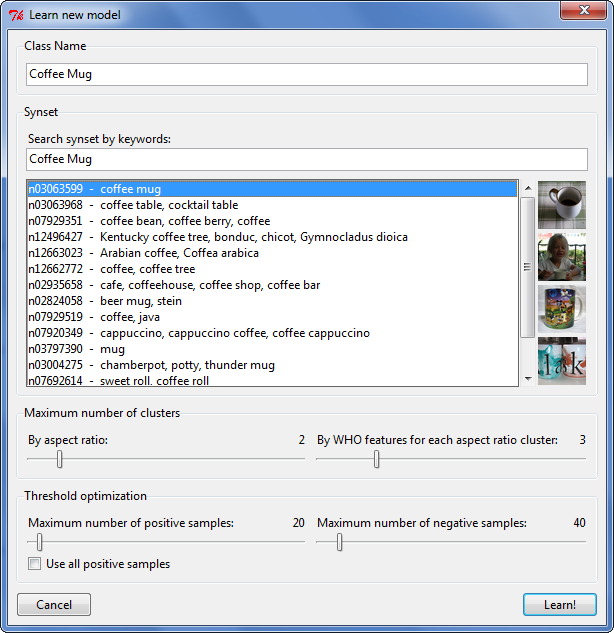}
    \caption{The \textit{Learn Model from ImageNet} dialog of the ARTOS GUI. It allows the user to select a synset and adjust some parameters related to clustering and threshold optimization.}
    \label{fig:learn_dialog}
\end{figure}

We use the \textit{ImageNet} dataset \citep{deng09} for automatic acquisition of a large set of samples for a specific object category. With more than 20,000 categories, \textit{ImageNet} is one of the largest non-properiatery image databases available. It provides an average of 300-500 images with bounding box annotations (annotated by crowd-sourcing) for more than 3,000 of those categories and, thus, is suitable for learning object detection models.
Everything a user has to do in order to learn a new model using the ARTOS GUI is to search for a synset and to click ``Learn!'' (see \figurename~\ref{fig:learn_dialog}).
For now, ARTOS requires access to a local copy of the \textit{ImageNet} images and annotations (or a subset at least), which must be available on the file system, but we are planning to change this in future with a download interface.

\begin{figure}[tbp]
    \centering
    \includegraphics[width=0.5\linewidth]{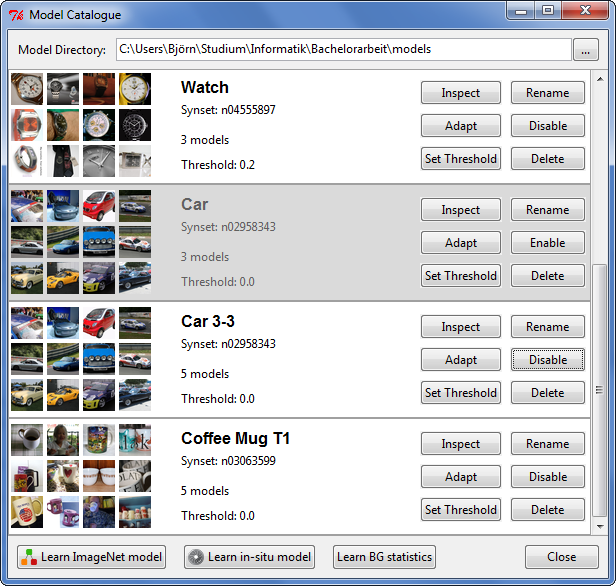}
    \caption{\textit{Model Catalogue} window of the GUI.}
    \label{fig:model_catalogue}
\end{figure}

\paragraph{Model Creation}

As feature representation, we use \textit{Histograms of Oriented Gradients} (HOG), originally proposed by \cite{dalal05}, with the modifications of \cite{felzenszwalb10} as features.
\cite{hariharan12} proposed a method for fast learning of models, even when only few positive and no negative samples are available. It is based on Linear Discriminant Analysis (LDA), which makes the following assumptions about class and feature probabilities:
\begin{align}
P[y] &= \varphi^y (1-\varphi)^{1-y} \\
P[\vec{x}|y] &\sim \mathcal{N}(\vec{\mu}_y, \mat{\Sigma})
\end{align}
From this, a linear classifier of the form $ w^Tx+b $ can be derived and $w$ turns out to be:
\begin{equation}
\label{WHO}
\vec{w} = \mat{\Sigma}^{-1}(\vec{\mu}_{\text{pos}} - \vec{\mu}_{\text{neg}})
\end{equation}
Important for a fast learning scheme is that $\vec{\mu}_{\text{neg}}$ and $\mat{\Sigma}$ do not depend on the positive samples and can be computed in advance and off-line.

In combination with HOG features, \cite{hariharan12} call the resulting features \textit{Whitened Histogram of Orientations} (WHO), although the ideas of \cite{hariharan12} can
be also used with other feature types that could be integrated into ARTOS.

ARTOS first performs two stages of clustering on the dataset obtained from \textit{ImageNet}: first, the images are divided into clusters by an aspect-ratio criterion, and then, each resulting cluster is subdivided with respect to the WHO features of the samples. This is done using a simple k-Means algorithm. One model is learned for each cluster according to equation \ref{WHO}. Those models are then combined in a model mixture for the object class.

\paragraph{Threshold Optimization}

While \cite{hariharan12} gave an explicit formula for $w$ in $w^Tx+b$, they kept quiet about how to obtain an appropriate bias $b$. To determine optimal biases, ARTOS finally runs a detector with the learned models on some of the positive samples and additional negative samples taken from other \textit{synsets} of \textit{ImageNet} in order to find a bias that maximizes the $ \text{F}_1\text{-Measure} $.

But finding the optimal threshold for each model of the mixture independently is not sufficient. Since the models are combined and the final detection score will be the maximum of the detection scores of the single models, an optimal combination of biases is crucial. Thus, we employ the heuristic \textit{Harmony Search} algorithm of \cite{geem01} to approximate an optimal bias combination that maximizes the $ \text{F}_1\text{-Measure} $ of the entire model. This could be easily adapted to other performance metrics or other optimization algorithms. In particular, we do not advocate for Harmony Search here and we believe that any other heuristic search method would work equally well.

\paragraph{Adaptation}

\begin{figure}[tbp]
    \centering
    \includegraphics[width=0.5\linewidth]{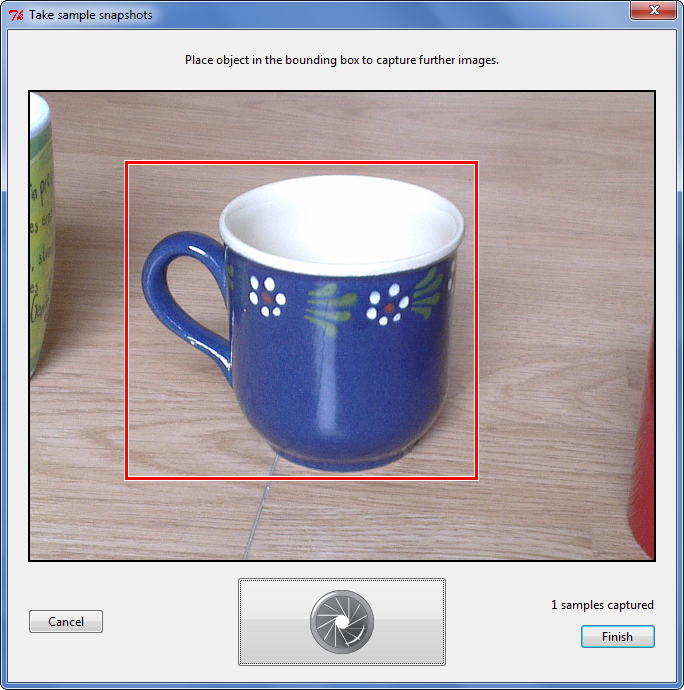}
    \caption{Models can be easily adapted by taking some in-situ images.}
    \label{fig:adaptation}
\end{figure}

After a model has been learned from \textit{ImageNet}, it can be adapted easily to overcome domain-shift effects. \textit{PyARTOS}, the Python based GUI to ARTOS, enables the user to take images using a camera (see \figurename~\ref{fig:adaptation}) or to annotate some image files, from which a new model will be learned and added to the model mixture.

\paragraph{Detection}

For fast and almost real-time object detection, ARTOS incorporates the FFLD library (\textit{Fast Fourier Linear Detector}) of \cite{ffld}, which leverages the \textit{Convolution Theorem} and some clever implementation techniques for fast template matching.

\section{Quantitative evaluation}

\begin{table}[tb]
    \centering
    \begin{tabular}{lc}
        \toprule
        method & mean average precision\\
        \midrule
        ImageNet model only (\textit{raptor}, \cite{Goehring14:ITR}) & $49.7\%$\\
        In-situ model only (\textit{raptor}, \cite{Goehring14:ITR}) & $36.6\%$\\
        Adapted/combined model (\textit{raptor}, \cite{Goehring14:ITR}) & $54.1\%$\\
        \midrule
        ImageNet model only (\textbf{\textit{artos}}) & $55.5\%$\\
        In-situ model only (\textbf{\textit{artos}}) & $51.0\%$\\
        Adapted/combined model (\textbf{\textit{artos}}) & $\mathbf{63.5\%}$\\
        \bottomrule
    \end{tabular}
    \caption{Results on the \textit{Office} dataset and a comparision to \cite{Goehring14:ITR}}
    \label{tab:office}
\end{table}

We followed the experimental setup of \cite{Goehring14:ITR} to evaluate ARTOS on the \textit{Office} dataset and we omit the details for the sake of brevity here.
The results given in Table~\ref{tab:office} reveal both the
benefit of adaptation as well as the general benefits of ARTOS. Both the clustering and the threshold optimization implemented in ARTOS contribute the performance benefit we observe here.

\section{How to get ARTOS and what are the next steps?}

A first (still not feature-complete) version of ARTOS has been released under the terms of the GNU GPL:
\begin{center}
\url{http://cvjena.github.io/artos/}
\end{center}
There is also a related \texttt{github} repository and we invite everyone to contribute and use our code for various vision applications.

We are planning to add a public model catalogue to the website of ARTOS so that people can 
upload and download models of common objects. The project is part of the lifelong learning initiative
of the computer vision group in Jena.

\vspace{16pt}
\begin{center}
\textbf{Enjoy object detection!}
\end{center}

\paragraph{Acknowledgements}
We would like to thank \cite{ffld} and \cite{hariharan12} for providing the source code of their research. Furthermore and most importantly, we also thank the authors of \cite{Goehring14:ITR}, who presented the approach on which our open source project is based on.

\vskip 0.2in
\bibliography{references}

\end{document}